\documentclass{article}
\usepackage{graphicx}
\usepackage{tcolorbox}
\usepackage{enumitem}
\usepackage{longtable}
\usepackage{lipsum}

\usepackage[main, final]{neurips_2026}

\usepackage[utf8]{inputenc} 
\usepackage[T1]{fontenc}    
\usepackage{hyperref}       
\usepackage{url}            
\usepackage{booktabs}       
\usepackage{amsfonts}       
\usepackage{nicefrac}       
\usepackage{microtype}      
\usepackage{xcolor}         

\usepackage{amsmath,amssymb}
\usepackage{multirow} 
\usepackage{array} 
\usepackage{adjustbox} 
\usepackage{colortbl}
\usepackage{makecell} 
\usepackage{subcaption}

\newcommand{\val}[2]{\makecell[l]{#1~{\tiny[#2]}}}

\title{Evaluating Multimodal Narrative Understanding of Popular Hollywood Films}

\author{%
  \textbf{David Bamman},$^1$ \textbf{Kent K. Chang},$^1$
  \textbf{Allison Cooper},$^2$
  \textbf{Juishan Hsu},$^3$
  \textbf{Reina Kushihashi},$^3$\\
  \textbf{Madison Mar},$^3$
  \textbf{Arnav Podichetty},$^3$
  \textbf{Rachael Samberg},$^4$
  \textbf{Ipek Nil Sancak}$^3$ and
  \textbf{Yuhan Shao}$^3$ \\[0.5em]
  $^1$School of Information, UC Berkeley \quad
  $^2$Cinema Studies, Bowdoin College \\
  $^3$UC Berkeley \quad
  $^4$Scholarly Communication and Information Policy, UC Berkeley \\[0.3em]
  \texttt{\{dbamman,kentkchang\}@berkeley.edu}
}

\begin{document}

\maketitle

\begin{abstract}
Multimodal language models increasingly show promise for enabling the large-scale computational analysis of film, opening up new avenues for learning about film history and the evolution of narrative techniques.  But the creation of stable benchmarks built around Hollywood films is complicated by copyright protections.  In this work, we address these concerns directly, by building a new collection of Hollywood films defined by two criteria: box office popularity (where we publish the first large-scale, open collection of weekly box office earnings reported by \emph{Variety} magazine from 1922--1979); and likely public domain status (by researching copyright registrations and renewals in the US \emph{Catalog of Copyright Entries}).  We build a new multimodal MCQ benchmark on top of this collection that focuses on narrative elements that directly evaluate the abilities of models to inform meaningful research on film narrative; we find that many vision-language models struggle on this task (with many performing at near-chance levels of accuracy), while audio-visual models (including those that use audio in captioning scenes) reach a maximum accuracy of 61.1\%, well below human-level performance.

\end{abstract}

\section{Introduction}

One of the most exciting consequences of advances in NLP, CV and AI has been in their application as analytical tools to shed light on questions of culture at scale. While this use has long focused on the domain of text\ \citep{underwood2019distant,piper2019enumerations}, we see increasing application for the analysis of film.  This work has shed light on changes in pacing and luminosity\ \citep{cutting2011quicker}, the representation of race and gender on screen\ \citep{guha2015gender,arnold2019visual,doi:10.1073/pnas.2409770121}, and comedic timing\ \citep{zribi2026timing}, along with many others\ \citep{9319168}.

However, while progress in other application areas of these methods can be driven by the formation of open benchmarks\ \citep{deng2009imagenet,lin2014microsoft,kay2017kinetics}, film as an object of study presents distinct challenges due to the copyrighted nature of the underlying materials.
While many of the core tasks---from shot boundary segmentation\ \citep{zabih1995feature,soucek2024transnet} to character identification\ \citep{everingham2006hello,Nagrani_2018_ECCV}---are often designed with application to the film industry in mind, restrictions on the digitization and republishing of copyrighted materials often leave researchers without direct access to the underlying data.  
Even if extracting clips for research purposes is a fair use—an exception to copyright owners' exclusive rights—film companies have a robust licensing market and often challenge uses that should be considered fair. Datasets that instead rely on practitioners downloading videos from URLs on YouTube also find themselves with increasingly more restricted access over time, as videos are removed by users or through compliance with Digital Millennium Copyright Act (``DMCA'') takedown requests.  
Benchmarks built on this data are unstable.

One solution to this challenge is to build benchmarks on movies in the public domain. The public domain includes not only works currently published before 1931 in the US,\footnote{Copyright protection extends for a fixed period of time. Films are typically works of corporate authorship protected for 95 years from publication; as of 2026, films published prior to 1931 are in the public domain. The matter is complicated, however, by the fact that elements of films may be remastered, and such changes to the original film would be protected by new copyright.} but also 
any film whose copyright owner (typically the production studio) did not renew its copyright during the period in US copyright law when such registration renewals were both available and required.
This, however, raises its own challenges: first, there is no known registry of public domain materials, and films that an online source might claim to be in the public domain can be contested; many of the feature films found on sources like the Internet Archive are still in copyright and could be subject to a DMCA takedown request.
Second, while thousands of movies are in the public domain, not all of them are equally notable---the public domain includes war propaganda released by the US government, films whose production studios failed to renew their copyright due to lack of interest in the film, and so on.  In building a benchmark around Hollywood films, we want to incorporate some measure of cultural significance as well.

In this work we address these critiques through two interventions in data sourcing: first, determining the cultural impression of films in the United States (measured by \emph{popularity} at the US box office); second, identifying films that are truly likely to be in the public domain---not through trusting the claims of a third party, but through researching that status with the US \emph{Catalog of Copyright Entries}.  These two criteria---cultural significance and openness---differentiate our work from related efforts to build datasets that include elements of Hollywood films\ \citep{tapaswi2016movieqa,rawal2024cinepile,wang2025lvbench}, including historical ones\ \citep{zaranis2025moviefactsfibsmf2}.

Given the dataset defined by these criteria, we build a benchmark around it for measuring the performance of multimodal models at the task of long-form \emph{narrative} understanding, focusing in particular on questions of temporality, plot, character, setting, perspective, representation, symbolism and object identification.  This benchmark lets us assess the performance of a range of models at measurement tasks increasingly driving scholarship in computational social science and the digital humanities.  
While previous benchmarks have explored long-form motion pictures\ \citep{zaranis2025moviefactsfibsmf2,wang2025lvbench} and interpretive tasks in other modalities like text\ \citep{sui-etal-2025-kristeva,hamilton-etal-2026-narrabench}, we bring these two paradigms together in this work to assess how such models can inform our analytical understanding of complex narrative phenomena in film.

Our work therefore makes the following contributions:

\begin{itemize}[leftmargin=* , itemsep=2pt]
    \item We present the first systematic database of historical box office earnings (extracted from \emph{Variety} magazine), spanning 1922--1979. This database is publicly available at \url{https://github.com/bamman-group/variety-boxoffice}.
    \item We present a new dataset of popular films that are likely in the public domain in the United States, which can form the basis for benchmarks that others can trust will be stable over time.
    \item We present a new benchmark for narrative understanding in these films, and assess the performance of several multimodal models at this difficult task. We find that many vision-language models struggle on this task (with many performing at near-chance levels of accuracy), while audio-visual models (including those that use audio in captioning scenes) reach a maximum performance of 61.1\%, well below human-level performance.  This Classical Hollywood Narrative Benchmark (and code to support it) is publicly available at \url{https://github.com/bamman-group/chnb}.
\end{itemize}

\section{Defining the collection}

\subsection{Identifying popular movies}

\begin{figure}
  \centering
  \begin{minipage}{0.45\textwidth}
    \centering
    \includegraphics[width=\textwidth]{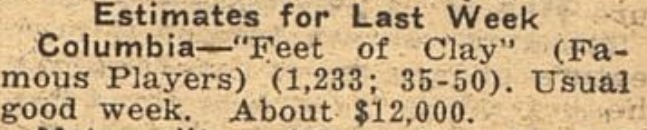}
  \end{minipage}
  \hfill
  \begin{minipage}{0.45\textwidth}
    \centering
    \includegraphics[width=\textwidth]{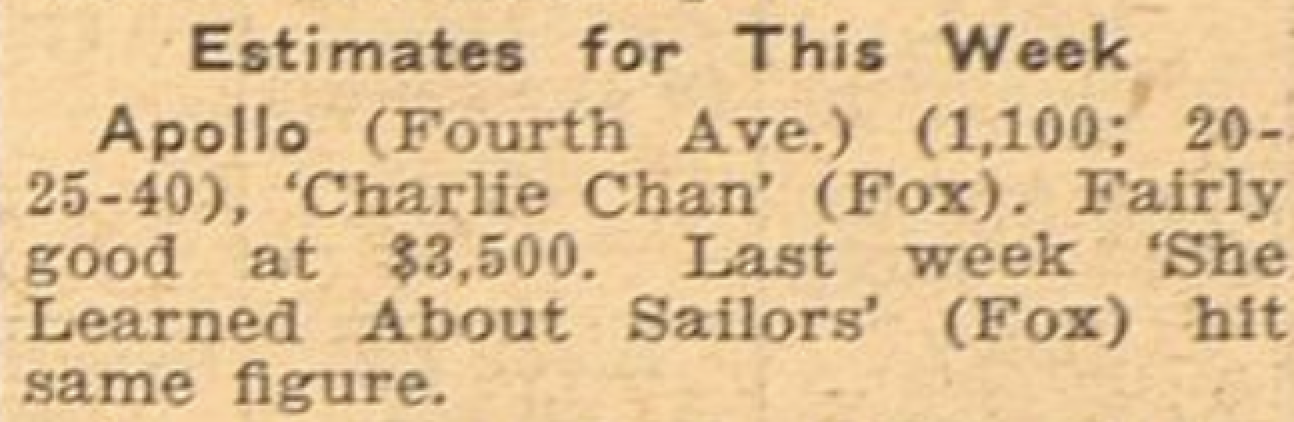}
  \end{minipage}
   \caption{Weekly box office information from two issues of \emph{Variety}, October 29, 1924 (left) and July 10, 1934 (right).}
   \label{variety}
\end{figure}

Our first goal is to identify movies that are popular. Historical box office data in the United States is fragmentary, with major sources of information coming from ledgers kept by executives at individual studios\ \citep{glancy1992mgm,jewell1994rko,glancy1995warner} and trade magazines such as \emph{Variety}, \emph{The Motion Picture Herald} and \emph{Hollywood Reporter}. \emph{Variety} has the deepest historical collection of box office information: starting with its March 3, 1922 issue (and persisting through the early 21st century), \emph{Variety} reports the weekly box office receipts for individual movies in specific theaters. Figure \ref{variety} gives two typical examples of this information.  

Within a broader column noting the city (Washington and Indianapolis), each entry lists the theater, movie title(s) and box office estimate for that week. 
We extract this information as a structured tuple---e.g., $\langle$ Washington, Columbia, \emph{Feet of Clay}, \$12,000 $\rangle$---from all pages of \emph{Variety} where the full text of at least one article on the page contains the phrase ``estimates for this week'', ``estimates for last week'', or where the title of an article contains the phrase ``picture grosses''. We prompt multimodal LLMs to extract all tuples from the scan of each page, providing two shots (input image, output json) to illustrate the desired behavior.

Movies mentioned in \emph{Variety} are often terse, relying on readers' general familiarity with movies that year (e.g., ``Gone'' to denote \emph{Gone with the Wind} in 1940). For each year, we manually create a mapping of aliases ({``Gone with the Wind'', ``Gone'', ``Gone with Wind'', etc.) to an IMDB record for that movie for all aliases appearing at least 3 times in a year.  This process allows us to generate a ranked list of movies each year by their total box office numbers reported by \emph{Variety}.

To evaluate the accuracy of this extraction method, we create a gold-standard dataset by manually labeling 4,072 tuples in 21 issues of \emph{Variety} spanning the years 1922--1979.  We use 7 of these issues for development and hold out 12 issues for held-out evaluation.  We primarily assess validity through the Spearman rank correlation coefficient between the resulting ranked lists of movies (comparing the ranks derived from human tuples vs. model predicted tuples), since the rank ultimately constitutes the decision criterion we use to define the collection below (the top 100 movies per year). We find that Gemini 3 Pro (with high thinking and ultrahigh image resolution) performs best on this held-out evaluation, with $\rho = 0.961$. The complete evaluation procedure can be found in Appendix \ref{boxofficeacc}.

We use this best performing model to extract this structured information for all \emph{Variety} issues from 1922--1979.  By aggregating the total box office earnings for all movies, we are able to assemble ranked lists of the top grossing movies each year over this complete time frame, comprising 1.4M weekly box office numbers for over 24,000 movies.  This represents, to our knowledge, the first comprehensive, openly available dataset of historical box office information for feature films.  All extractions, alias mapping, and aggregated yearly/weekly charts are openly available at \url{https://github.com/bamman-group/variety-boxoffice}.

These top charts allow us to define a measure of popularity, including a film within the boundaries of our dataset if it is among the top 100 highest-grossing films in a year.
Commercially distributed features form a logical corpus for a benchmark focusing on narrative understanding because they were produced in industrial conditions that prioritized narrative coherence and broad legibility---the properties we seek to evaluate. The \textit{Variety }dataset is rich in films from the Classical Hollywood era, broadly defined as the period spanning 1917--1960 \citep{bordwell1985classical} and marked by industrial and aesthetic developments such as the studio system and the continuity system, created to improve the clarity and efficiency of storytelling onscreen. Other dominant narrative trends of the period include character-centered causality, goal-oriented plots, and an emphasis on narrative closure \citep{bordwell1985narration}, along with the evolution of the star persona \citep{dyer1998stars}, strategies to navigate the Hollywood Production Code \citep{jacobs1997wages}, and the formation and reinforcement of genre conventions \citep{altman1999film}.  The dataset spans several years of the Post-Classical period as well;
 this era saw a loosening of character-centered causality \citep{elsaesser1975pathos}, increasingly ambiguous endings \citep{bordwell2006way}, the reworking of Classical-era genres \citep{ray2020certain}, and the growing influence of European art cinema \citep{bordwell1979art}. 

Finally, we note that the limitations of using box office as a proxy for significance are well documented \citep{maltby2006prospect, staiger1992interpreting}. 
Film scholars have cautioned against an over-reliance on commercial metrics, arguing that too narrow of an approach misses aesthetically important works \citep{johnston1973women, rich2013new} or the broader historical and social context of cinema \citep{allen1985film, verhoeven2019re}. Using box office gross as the main criterion in assembling our collection of popular films means our corpus excludes films that circulated outside mainstream distribution in the United States, from the ``race'' films of the silent era \citep{stewart2005migrating} to transnational festival films \citep{white2015women}. It is  worth emphasizing that box office is but one of many measures of significance that might be adopted to assemble a film dataset to evaluate multimodal understanding of narrative. Our methodology for identifying the public domain status of popular films, described in detail in the following section, can easily be adapted for the development of benchmarks with different boundaries. 

\subsection{Investigating public domain status}
\label{pdstatus}

In the United States, most movies published prior to 1931 have fallen into the public domain (as of the time of this writing); as have most movies whose copyright was registered prior to 1964 but failed to be renewed 28 years later. While the first criterion allows us to identify likely public domain movies relatively easily by considering their year of publication, the latter criterion is more difficult since it requires a.) identifying the year of copyright registration for a film and b.) identifying its \emph{lack} of renewal (including the renewal of any screenplays or expressive works from which the film was adapted, which may bear separate registration).\footnote{Copyright renewal matters only for a certain time periods. Beginning with the Copyright Act of 1976 (effective January 1, 1978), neither registration nor renewal is required for films to be protected by copyright. Prior to the 1976 Act, however, there is a complex landscape of protection based on a combination of authorship (individual vs. corporate), publication status, registration date, and renewal.}

We turn to two sources for identifying this information: the \emph{Catalog of Copyright Entries}, published by the US Copyright Office through 1978; and the Copyright Public Records System (CPRS), an online database published by the US Copyright Office, which contains registrations (and renewals) from 1978 forward. To identify candidate movies that may be in the public domain for lack of registration renewal, we extracted all registrations and renewals for movies recorded in the print \emph{Catalog of Copyright Entries} using digitized versions on the Internet Archive (which captures registrations/renewals until 1978) and extracted all motion picture renewals from the Copyright Public Records System to capture renewals made after 1978.

Using this information, we matched movies between the \textsc{Registered} set and the \textsc{Renewed} set using the original registration number (which often appears as a canonical identifier in both sets); any movie that was matched was automatically excluded, since this provides evidence that the movie was likely appropriately renewed.  Since registration numbers may change---whether through OCR mistakes, mistyping, or other factors---we also carried out a detailed manual review of the movies that did not match, attempting to match them based on the similarity of their title.  

We also check for one other situation that would lead a movie that has not been renewed to still be under copyright: even if the copyright for the film was not renewed, if the source (such as a short story or novel) was appropriately copyrighted and renewed, then the underlying story for the movie may still remain in copyright as well (e.g., as is the case for \emph{It's a Wonderful Life}). We draw on data from the American Film Institute, which provides information about whether a movie was based on some other original (e.g., literary) source; any movie described by AFI as being based on an additional source in copyright was removed from the collection.

The two criteria laid out above---popularity (among the top 100 movies per year by box office revenues) and likely public domain status---provide the conceptual boundaries for this collection.  We adopt these stringent criteria in order to minimize DMCA takedown requests, and source the content of the films themselves from the Internet Archive, further limiting the collection to only sound films (i.e., no silent-era movies). This results in dataset of 61 popular movies.

\section{Building a benchmark}

Given this collection of popular films whose copyright status are unlikely to be challenged, we build a benchmark around it to assess the long-form narrative understanding capabilities of multimodal models. We focus in particular on questions that illustrate the affordances of such models for work in cultural analytics of film.  For ease of evaluation, we frame the task as a multiple-choice question format, with four answer options (only one of which is correct).
We focus on eight narrative categories described below:

\begin{itemize}[leftmargin=*]
    \item \textbf{Temporality.} Questions that track attention to the order of events---both as they occur chronologically within the story world and as they are depicted to the viewer; these orderings are in tension in cases of anachrony \citep{genette1980narrative}, such as flashbacks and flashforwards. \\
    \emph{Example question:} What is the order in which the four characters are arrested? A.) Countess de Mavon $\rightarrow$ Nurse Edith Cavell $\rightarrow$ Mme. Moulin $\rightarrow$ Mme. Rappard. B.) \ldots

    \item \textbf{Plot.} Questions that identify the narrative function of a scene, whether it introduces a complication, raises the stakes, resolves the central tension, and to track whether character goals stated early in the film are ultimately fulfilled. \\
    \emph{Example question:} Why didn't Charles leave the crime scene right away after murdering Meinike? A.) He is setting up a paper trail; B.) \ldots

    \item \textbf{Character.} Questions that identify roles and track character dynamics based on what the film shows: what characters do, say, how they are dressed, and how they are staged relative to one another. \\
    \emph{Example question:} Which character is shown smoking? A.) Jack; B.) \ldots

    \item \textbf{Setting.} Questions that identify and distinguish between locations in the film, and observe how the physical staging of characters within a space (e.g.\ social blocking) conveys power, relationship, and intention, which require attention to mise-en-sc\`ene rather than plot. \\
    \emph{Example question:} Which of these locations do we not see the interior of? A.) King Little’s castle; B.) \ldots

    \item \textbf{Perspective.} Questions that identify from whose vantage point events are presented, and whether the film ever gives the audience information the characters themselves do not have. This tests attention to how the film is narrated---not what happens, but who knows what, and when. \\
    \emph{Example question:} When Norma tells Michael how much she loves him, who sees Dr.\ Besant enter the room first? A.) Norma; B.) We do as viewers (before any characters); C.) \ldots

    \item \textbf{Representation.} Questions that identify how the film depicts gender roles, social identity, and group membership, based on what is shown and said in the film. \\
    \emph{Example question:} Does this movie pass the Bechdel test? (Two named women talking to each other about a topic that is not a man.) A.) Yes, between Irma and Phyllis repeatedly \ldots B.) \ldots

    \item \textbf{Symbolism.} Questions that identify objects, sounds, or visual actions that recur across the film and carry narrative or thematic weight. \\
    \emph{Example question:} In Marilyn's performance where she is surrounded by dishes that she is washing, what does this chore symbolize, based on the lyrics to the song she sings? A.) Growing up; B.) \ldots

    \item \textbf{Object identification.} Questions that identify a specific on-screen object, track where it appears or what is done with it, or connect it to its function in the plot. \\
    \emph{Example question:} What object is repeatedly used by neighbors to cope with the heat? A.) Hand fans; B.) \ldots
\end{itemize}

\begin{figure}[t]
\centering

\begin{subfigure}{0.325\textwidth}
  \includegraphics[width=\linewidth]{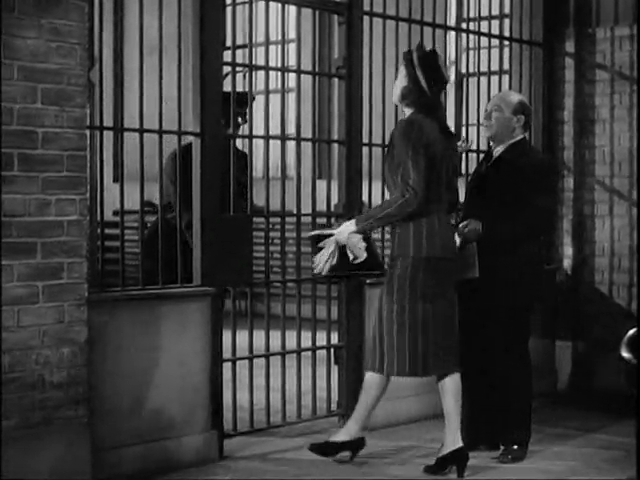}
\end{subfigure}
\begin{subfigure}{0.325\textwidth}
  \includegraphics[width=\linewidth]{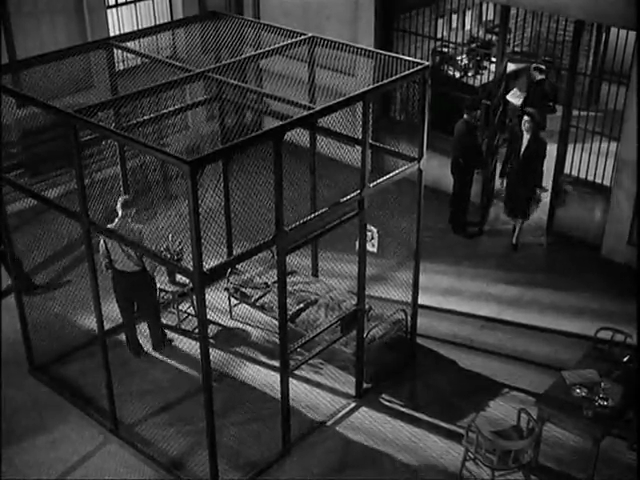}
\end{subfigure}
\begin{subfigure}{0.325\textwidth}
  \includegraphics[width=\linewidth]{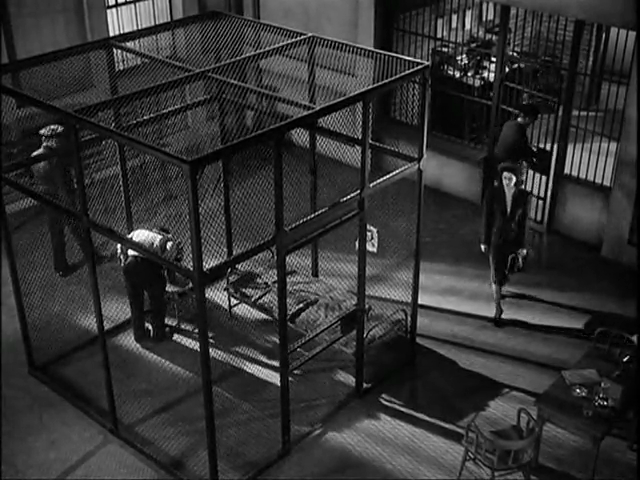}
\end{subfigure}

\caption{Example question (\emph{His Girl Friday}): ``How is the camera positioned as Hildy enters the room to talk to Earl Williams?'' \emph{A}.) On Hildy's side of the bars to look through the grate at Earl. \emph{B}.) The camera is at eye level and moves parallel to her. \emph{C}.) On the inside of the cell to look out at Hildy.	\emph{D}.) High angle above Hildy and Earl.}
\end{figure}

To create benchmark questions, seven co-authors viewed the entirety of a movie and created an average of 12.8 questions for each one, resulting in an initial set of 779 questions across 61 films. We use a plain-language scene description to refer to specific scenes (not explicit timestamps) and avoid distractors that are obviously off-topic to avoid simply testing commonsense reasoning capability.  We assess expert-level human performance by distributing questions for a sample of ten movies (133 questions) to co-authors who did not write those questions, asking them to watch the movie and answer all questions (going back and forth to the movie as needed); we find human-level accuracy to reach $82.0\%$, reflecting in part the complex nature of narrative inferences.  Sources of error include ambiguity in the question/answer options (where multiple choices could be argued to be correct), but also reflects the natural difficulty of some information-seeking questions (e.g., where  attention is required to a scene that is easily missed).  Table \ref{memcat} (Appendix \ref{memorization}) lists the distribution of annotated categories, with greatest representation of questions around plot, character and setting.

\section{Memorization}

One of the challenges of working with popular movies is that they are frequently discussed online, and these discussions make their way into the pre-training data for LLMs.  Past work has found this to be the case as well: \cite{zaranis2025moviefactsfibsmf2} report an accuracy of $66.3\%$ (compared to a random performance of $50\%$) when prompting Gemini 2.5 Pro to answer questions based on the movie title and date of release alone (with no access to the video); \citet{asadi2026mirage} find this kind of ``mirage reasoning'' prevalent in multimodal medical benchmarks.  We see this as an example of test data contamination \citep{dodge2021documenting, chang-etal-2023-speak}, where models use metapragmatic information \emph{about} a movie---rather than the content of the movie itself---to make decisions.

To account for this, we pass all questions through three frontier LLMs---Gemini Pro 3.1, Claude Opus 4.7 and GPT 5.5---with the following prompt: ``Based on your knowledge of the movie \{\texttt{MOVIE}\} (\{\texttt{YEAR}\}), answer the following question.''  All models exhibit similar rates of memorization (Gemini $39.4\%$, Opus $40.7\%$, GPT $38.8\%$).  To mitigate this effect, we subselect questions from the pool so that the performance across all models when prompted with the movie title and date alone is approximately $25\%$ (reflecting a random guess), detailed in Appendix \ref{memorization}.   
This yields a total of 628 benchmark questions (discarding 151 from the original pool).  As Table \ref{memcat} (Appendix \ref{memorization}) illustrates, the exclusion rate varies by category: models have internal knowledge of common analytical discussion topics about the film---including representation and symbolism---and much less so about questions that require access to specifics visuals.
For convenience, we let $\mathcal{B}$ denote the post-filter benchmark of 628 questions used in subsequent experiments.

\section{Experiments}
\label{sec:experiments}

\subsection{Setup}

We evaluate five paradigms on $\mathcal{B}$. 
For a film with video $V$ (frames and audio) and a question $q$, each paradigm involves a model $M$ that predicts an answer $\hat{a} = M(c, q)$ from a context $c$ that varies by how $V$ is compressed:

\paragraph{Closed-book baseline: $c=\varnothing$.}
The system answers from $q$ and 
its parametric knowledge alone.
This establishes whether the benchmark is solvable without the movie at all, a precondition for any subsequent gain to be attributable to movie content rather than priors (visual and textual).

\paragraph{Subtitles-only baseline: $c=\mathcal{S}$.}  
The QA backbone takes only the dialogue transcript $\mathcal{S}$ and the question, without visual or frame-derived input. $\mathcal{S}$ is generated by transcribing the audio track using Distil-Whisper \texttt{large-v3}\ \citep{gandhi2023distilwhisper}.
Prior work on long-movie comprehension has found subtitle-only access to be a surprisingly strong baseline~\citep{zaranis2025moviefactsfibsmf2}.
We include it as a baseline both for performance comparison and to isolate dialogue as a separate point for long-video compression strategies for narrative understanding.

\paragraph{End-to-end: $c\subseteq V$.} 
The QA model is itself a multimodal model and extracts a fixed sample of V directly.
We consider the following types of end-to-end models:
a.) \textit{Long-video models} are architectures designed for long-form video and extracts a uniform 64-frame sample of the film.
b.) \textit{Vision--language models} are general-purpose VLMs and take a uniform 256-frame sample.
c.) \textit{The audio-visual model} (Gemini~3~Flash) processes the video in its entirety, including the audio track.

\paragraph{Socratic~\citep{Zeng2023-lh}: $c=$\textsc{Captioner}$(V)$.}  
We follow the two-stage pipeline described in~\cite{NEURIPS2024_5f280960}:
First, a \textsc{Captioner} produces one of the following:
a.) \textit{Frame captions} are generated from a 0.5-fps frame strip with no audio; the captioner sees roughly thirty stills per minute of film.
b.) \textit{Clip captions} are generated from video chunks (target duration 60 seconds) that respect shot boundaries~\citep{soucek2024transnet} and include the audio track.
These captions are then timestamped and concatenated chronologically into a world state history~\citep{NEURIPS2024_5f280960} that the QA backbone receives (instead of $V$). 
This class tests whether textual compression preserves the audio-visual signal for long-video QA.
We adopt Gemini 3 Flash as the captioner.

\paragraph{Agentic retrieval: $c=\pi(V, q)$.}
To test whether query-conditioned selectivity improves on the Socratic baseline (which has access to full content), we instantiate the retrieval policy $\pi$ in two ways:
a.) \textit{Frame image retrieval} is the VideoAgent loop described in~\cite{fan2025videoagent}: $\pi$ starts from 8 uniform frames in a 64-frame pool and re-fetches up to four more per iteration via CLIP~\citep{pmlr-v139-radford21a} cosine similarity on a confidence-threshold loop ($\leq3$ iterations).
b.) \textit{Caption retrieval} utilizes the Letta agent~\citep{Packer2023-rt}, where $\pi$ loads the Socratic captions (both frame- and clip-based) into archival memory and lets the QA backbone decide to trigger a semantic search.\footnote{\url{https://docs.letta.com/api/python/resources/agents/subresources/passages/methods/search/}. We use the default \texttt{text-embedding-3-small} model for semantic search.}

\paragraph{Pre-processing pipeline.}
Several film-level artifacts are precomputed once per film and shared across paradigms:
a.) Uniform frame samples at $N \in \{64, 128, 256\}$ are extracted at indices $\lfloor i \cdot T / N \rfloor$ where $T$ is the film's frame count, and cached as JPEGs.
b.) Shot-grouped chunks come from detected shot boundaries merged greedily to a target duration of 60 seconds, with a 10-second minimum and the constraint that no shot is split.
c.) Per-frame CLIP embeddings of the largest uniform pool, $L_2$-normalized and cached, drive the targeted frame-retrieval step in the VideoAgent loop.
d.) Captions are produced by Gemini~3~Flash from either the 0.5-fps frame strip (frame captions) or the shot-grouped video chunk with audio (clip captions).

\begin{table}[t]
\centering
\small
\setlength{\tabcolsep}{5pt}
\renewcommand{\arraystretch}{0.9}

\caption{Accuracy on $\mathcal{B}$. The largest 95\% Wald confidence interval is $\pm 3.9\%$.  Bold indicates best overall; \underline{underline} marks the within-subgroup leader significantly better than the runner-up (whose 95\% CIs do not overlap).}

\label{tab:acc-final-aligned}

\begin{tabular}{@{}c@{\hspace{2.1em}}c@{}}

\begin{tabular}[t]{lcc}
\toprule
\textbf{Model} & \textbf{Compute} & \textbf{Acc.} \\
\midrule

\multicolumn{3}{c}{\textit{Closed-book} (question only)} \\
\midrule
Qwen3-VL-8B             & $\approx$0h 07m & 21.5 \\
GLM-4.1V-9B-Thinking    & $\approx$2h 46m & 20.4 \\
GPT-5-mini              & $\approx$0h 03m & 25.2 \\
Claude Haiku 4.5        & $\approx$0h 09m & 22.1 \\
Gemini 3 Flash          & $\approx$2h 47m & {26.6} \\
\midrule

\multicolumn{3}{c}{\textit{Subtitles only}} \\
\midrule
Qwen3-VL-8B             & $\approx$0h 20m & 32.2 \\
GLM-4.1V-9B-Thinking    & $\approx$31h 21m & 27.7 \\
GPT-5-mini              & $\approx$9h 22m  & 39.2 \\
Claude Haiku 4.5        & $\approx$0h 08m  & 36.9 \\
Gemini 3 Flash          & $\approx$2h 04m  & \underline{48.1} \\
\midrule

\multicolumn{3}{c}{\textit{End-to-end}} \\
\midrule
\multicolumn{3}{>{\columncolor[gray]{0.92}}c}{\textsc{\MakeLowercase{Long-video}} (64 frames)}\\
\midrule
VAMBA-Qwen2-VL-7B       & $\approx$1h 47m & 23.7 \\
VideoChat-Flash         & $\approx$1h 04m & {27.2} \\
HourLLaVA               & $\approx$1h 02m & 23.4 \\
LLaVA-NeXT-Video-7B-DPO & $\approx$0h 15m & 25.0 \\
\midrule

\multicolumn{3}{>{\columncolor[gray]{0.92}}c}{\textsc{\MakeLowercase{Vision--language}} (256 frames)}\\
\midrule
Qwen3-VL-8B      & $\approx$4h 43m & 29.1 \\
GLM-4.1V-9B-Thinking    & $\approx$6h 54m & 24.2 \\
GPT-5-mini              & $\approx$7h 52m & 34.2 \\
Gemini 3 Flash          & $\approx$6h 22m & {39.3} \\
\midrule

\multicolumn{3}{>{\columncolor[gray]{0.92}}c}{\textsc{\MakeLowercase{Audio-visual}} (full video)}\\
\midrule
Gemini 3 Flash  & $\approx$37h 24m & \textbf{61.1} \\
\bottomrule
\end{tabular}

&

\begin{tabular}[t]{lcc}
\toprule
\textbf{Model} & \textbf{Compute} & \textbf{Acc.} \\
\midrule

\multicolumn{3}{c}{\textit{Socratic}} \\
\midrule
\multicolumn{3}{>{\columncolor[gray]{0.92}}c}{\textsc{\MakeLowercase{Frame-based}}}\\
\midrule
Qwen3-VL-8B             & $\approx$1h 36m & 30.3 \\
GPT-5-mini              & $\approx$4h 42m & 36.8 \\
Claude Haiku 4.5        & $\approx$0h 58m & 35.0 \\
Gemini 3 Flash          & $\approx$7h 15m & \underline{46.0} \\
\midrule

\multicolumn{3}{>{\columncolor[gray]{0.92}}c}{\textsc{\MakeLowercase{Clip-based}}}\\
\midrule
Qwen3-VL-8B             & $\approx$1h 44m & 32.8 \\
GPT-5-mini              & $\approx$3h 30m & 48.1 \\
Claude Haiku 4.5        & $\approx$1h 02m & 45.9 \\
Gemini 3 Flash          & $\approx$2h 53m & \underline{58.9} \\
\midrule

\multicolumn{3}{c}{\textit{Agentic retrieval}} \\
\midrule
\multicolumn{3}{>{\columncolor[gray]{0.92}}c}{\textsc{\MakeLowercase{Frame image}} (VideoAgent)}\\
\midrule
Qwen3-VL-8B             & $\approx$16h 00m & 22.9 \\
GPT-5-mini              & $\approx$9h 22m  & 24.8 \\
Claude Haiku 4.5        & $\approx$1h 53m  & 26.1 \\
Gemini 3 Flash          & $\approx$2h 04m  & 28.8 \\
\midrule

\multicolumn{3}{>{\columncolor[gray]{0.92}}c}{\textsc{\MakeLowercase{Frame caption}} (Letta)}\\
\midrule
Qwen3-VL-8B             & $\approx$1h 31m & 26.1 \\
GPT-5-mini              & $\approx$1h 57m & 33.6 \\
Claude Haiku 4.5        & $\approx$3h 27m & \underline{39.8} \\
Gemini 3 Flash          & $\approx$4h 48m & 26.6 \\
\midrule

\multicolumn{3}{>{\columncolor[gray]{0.92}}c}{\textsc{\MakeLowercase{Clip caption}} (Letta)}\\
\midrule
Qwen3-VL-8B             & $\approx$1h 50m & 26.8 \\
GPT-5-mini              & $\approx$2h 13m & 26.0 \\
Claude Haiku 4.5        & $\approx$1h 51m & \underline{41.2} \\
Gemini 3 Flash          & $\approx$6h 51m & 30.3 \\
\bottomrule
\end{tabular}

\end{tabular}

\end{table}

\subsection{Results}
\label{sec:experiments-results}

We evaluate $\mathcal{B}$ on the following:~
HourLLaVA~\citep{NEURIPS2025_19741c61}, VAMBA-Qwen2-VL-7B~\citep{Ren_2025_ICCV}, VideoChat-Flash~\citep{li2026videochatflash}, and LLaVA-NeXT-Video-DPO~\citep{zhang2024llavanext-video} are long-video models; Qwen3-VL-8B~\citep{Qwen3-VL} and GLM-4.1V-9B-Thinking~\citep{Team2026-jr} are open-weight VLMs; and finally, closed-source models: GPT-5-mini~\citep{Singh2025-ag}, Claude Haiku 4.5~\citep{Anthropic2025-ko}, and Gemini~3~Flash~\citep{Gemini-Team2023-oj}.\footnote{Gemini 3 models are accessed via Vertex AI:~\url{https://docs.cloud.google.com/vertex-ai/generative-ai/docs/models}.}
Table \ref{tab:acc-final-aligned} summarizes model performance.
For each setup, we report accuracy and note the widest 95\% Wald confidence intervals to facilitate testing the significance of direct model comparisons.

In the closed-book baseline, no lower CI bound exceeds 25\%; models cannot answer questions in $\mathcal{B}$ from their parametric knowledge alone.
Nor do the closed-source backbones agree on which films they know better: per-film closed-book accuracies correlate at $\rho = +0.11$ ($p = 0.40$) between Gemini~Flash~3 and GPT-5-mini, $+0.28$ ($p=0.027$) between Gemini~Flash~3 and Claude Haiku 4.5, and $+0.44$ ($p < 0.001$) between GPT-5-mini and Claude.
The subtitle-only baseline shows that the questions are meaningfully answerable from real movie content; dialogue alone (the ASR transcript) raises accuracy to $48.1$ for Gemini, $39.2$ for GPT, and $36.9$ for Claude.

\paragraph{Clip-based Socratic captioning is competitive with native long-video processing.} 
The strongest end-to-end configuration is Gemini~3~Flash on full video ($61.1$ $\pm 3.8$), followed by the same model on clip-based captions  ($58.9$ $\pm 3.8$). 
Given the overlapping CIs, there is no meaningful gap between the text-mediated compression via clip captioning and Gemini's native multimodal processing.
Clip-based Socratic captioning is effective across models, which makes an empirical case for caption-based compression as a viable alternative to video processing.

\paragraph{Performance of agentic and Socratic methods differs across QA backbones.} 
The Letta agent and Socratic pipelines take the same captions but differ in how they reach the model: Letta retrieves from archival memory, but Socratic includes the entire chronological description based on the captions in the prompt. 
The Letta--Socratic gap is significant for Gemini on both caption types and for GPT on clip captions; for Claude the two paradigms are tied, and it is the strongest agentic-retrieval backbone of the three.
In contrast, frame-image retrieval barely exceeds chance on every backbone.
The agent's iterative frame-retrieval loop cannot compensate for what its modality omits (audio and dialogue).
Its CLIP-similarity is unsuitable for retrieving content in the audio stream that is relevant to the answer.

\paragraph{Frame budget exerts limited impact.}
For the six end-to-end backbones run at multiple frame budgets, we report average accuracy in Table~\ref{tab:frame-sweep} (Appendix \ref{framebudget}) for $N \in \{64, 128, 256\}$.
The largest within-backbone gain is $+3.6$ pp from $N{=}64$ to $N{=}256$; within-backbone CIs overlap heavily across the three budgets, and we therefore use a single canonical budget per sub-paradigm in Table~\ref{tab:acc-final-aligned} ($N{=}64$ for long-video models, $N{=}256$ for vision--language models).

\paragraph{Vision is not sufficient for film narrative understanding.}
Of the four architectures designed for hour-scale video at 64 frames (HourLLaVA, VAMBA-Qwen2-VL-7B, and VideoChat-Flash, LLaVA-NeXT-Video-DPO), none are statistically indistinguishable from the closed-book baseline.
The general-purpose vision--language models with a larger frame budget do better, but in the case of Gemini~3~Flash, switching from a 256-frame visual input ($39.3\%$) to full video, including audio track ($61.1\%$), gains $+21.8$ pp.
We observe similar patterns inside Socratic: holding the QA backbone fixed, swapping the captions from frame-based (no audio) to clip-based introduces significant gains for all three models.

\paragraph{Performance gains come from dialogue access and reasoning over video content.}
Per-film accuracy in each backbone's best configuration correlates positively with that backbone's subtitles-only accuracy (Table~\ref{tab:perf-corr}; $\rho_{\text{subtitles}}$ ranges from $+0.40$ to $+0.52$, all $p \le 0.002$): the films where the strongest configurations win are in the films where dialogue alone is informative.   
To assess whether memorization drives the performance gains on $\mathcal{B}$, we measure two proxies: web prevalence and title-prediction accuracy.
Web prevalence is the $\log_{10}(1+h_v)$-scaled count of Google Search results ($h_v$) for the title query, and title-prediction accuracy captures the fraction of 10 uniformly-spaced 5-frame window per film from which the backbone correctly predicts the canonical film title on IMDb, assessed by case-insensitive exact string match ($28.4\%$ for Gemini 3 Flash, $3.3\%$ for GPT-5-mini, and $2.1\%$ for Claude Haiku 4.5).
A model whose performance partly relies on memorized content would perform better on films with greater web presence and whose visual iconography it can identify.
However, in Table~\ref{tab:perf-corr}, we see that across all three closed models, $\rho_{\text{hits}}$ and $\rho_{\text{title}}$ are negative, suggesting films that are popular online and visually recognizable by the model are not easier even on the strongest configurations.

\begin{table}[t]
  \caption{Spearman rank correlations across all 61 films, of per-film accuracy in each backbone's best-performing configuration, compared with four predictors: closed-book accuracy of Gemini 3 Flash, subtitles-only accuracy, web prevalence, and title prediction accuracy.}
  \label{tab:perf-corr}
  \bigskip
  \centering
  \footnotesize
  \setlength{\tabcolsep}{5pt}
  \renewcommand{\arraystretch}{0.95}
  \begin{adjustbox}{max width=\textwidth}
    \begin{tabular}{l l c r r r r r r r r}
    \toprule
    \multirow{2}{*}{Model} & \multirow{2}{*}{Best} & \multirow{2}{*}{\textbf{Acc.}}
      & \multicolumn{2}{c}{Gemini closed-book}
      & \multicolumn{2}{c}{Subtitles}
      & \multicolumn{2}{c}{Web hits}
      & \multicolumn{2}{c}{Title prediction} \\
    \cmidrule(lr){4-5} \cmidrule(lr){6-7} \cmidrule(lr){8-9} \cmidrule(lr){10-11}
      & & & $\rho_{\text{gcb}}$ & $p$ & $\rho_{\text{subtitles}}$ & $p$ & $\rho_{\text{hits}}$ & $p$ & $\rho_{\text{title}}$ & $p$ \\
    \midrule
    GPT-5-mini       & Socratic, clip captions & $\mathbf{48.1}$ & $+0.22$ & $0.084$       & $+0.40$ & $0.002$       & $-0.25$ & $0.06$       & $-0.13$ & $0.31$ \\
    Claude Haiku 4.5 & Socratic, clip captions & $\mathbf{45.9}$ & $+0.37$ & $0.003$      & $+0.52$ & ${<}0.001$  & $-0.45$ & ${<}0.001$ & $-0.16$ & $0.23$ \\
    Gemini 3 Flash   & end-to-end, full video  & $\mathbf{61.1}$ & $+0.43$ & ${<}0.001$ & $+0.42$ & ${<}0.001$  & $-0.27$ & $0.04$       & $-0.26$ & $0.05$ \\
    \bottomrule
    \end{tabular}
  \end{adjustbox}
\end{table}

Gemini's $\rho_{\text{gcb}}=+0.43$ is a within-model consistency effect, but for Claude and GPT-5-mini, their clip-captioned Socratic routes the audio-visual content through Gemini Flash as a captioner before reaching the QA backbone. 
The $+0.37$ and $+0.22$ we observe for those two against the closed-book ranking of Gemini Flash, then, show that Gemini's parametric film knowledge influences its caption output enough to leave a per-film signal in the performance of the downstream models.

\section{Conclusion}

We present in this work a new benchmark of narrative questions built around popular movies from the Classical Hollywood era, defining the boundaries of that collection by movies that are popular (as measured by box office numbers reported by \emph{Variety} magazine) and whose copyright status is unlikely to be challenged (either by being released prior to 1931 or by registering their copyright but failing to renew it).  The questions require attention to complex narrative elements involving plot, character, setting, perspective, and more, and prove challenging for frontier multimodal language models.  As more research leverages such models for the large-scale computational analysis of film, we expect this benchmark to provide a proving ground for assessing comparative model performance.  Data and code to support this work are available at~\url{https://github.com/bamman-group/variety-boxoffice} and~\url{https://github.com/bamman-group/chnb}.

\section{Limitations}

While this work aims to address a gap in long-form multimodal benchmarks for assessing narrative understanding abilities of contemporary models, it is limited in several ways. By selecting movies based on popularity alone, we encode only one of the many possible forms of cultural significance, and omit movies from the time period that circulated outside of major metropolitan cities in the United States. The process we describe for investigating public domain status, however, could be applied to define new collections under alternative criteria.   
Additionally, all questions in the benchmark were created by researchers (of varying disciplinary backgrounds) at U.S. universities, which influences the narrative aspects in a film we find salient.  Finally, the benchmark specifically covers the era of Classical Hollywood cinema (through 1963); while we expect models with good long-form narrative understanding to be able to perform well on this data, performance may not generalize to films outside of this time period (both older and newer); we see this as a necessary trade-off for defining a collection of movies on which additional stable benchmarks can be built.

\section*{Acknowledgments}

The research reported in this article was supported by the Humanities and AI Virtual Institute (HAVI), a program of Schmidt Sciences, and by Google. This research used the Savio computational cluster resource provided by the Berkeley Research Computing program at the University of California, Berkeley.

\bibliographystyle{plainnat}
\bibliography{references}


\appendix

\section{Author contributions}

\begin{itemize}
    \item Conceptualization of narrative categories: DB, KC, AC, JH, RK, MM, AP, INS, YS

 \item  Benchmark question annotation: JH, RK, MM, AP, INS, YS (major); DB (minor)

 \item  Benchmark answer annotation (evaluation): DB, KC, AC, JH, RK, MM, AP, RS, INS, YS

 \item \emph{Variety} box office annotation: MM, DB

 \item Data curation (\emph{Variety}, \emph{Catalog of Copyright Entries}, Internet Archive): DB

 \item Statistical analysis: DB, KC

 \item Computational methodology: KC, DB

 \item Validation: DB, KC

 \item Writing: DB, AC, KC, RS

\end{itemize}

\section{Box office extraction accuracy}
\label{boxofficeacc}

Our procedure for generating ranked lists of movies per year by their total box office numbers involves using a multimodal LLM to extract individual $\langle$ city, theater, movie, gross $\rangle$ tuples from pages of \emph{Variety} magazine, and map movie titles to IMDB identifiers (as described in the main body above).  We evaluate the accuracy of this process by comparing the extractions derived from several models with those manually created by visually inspecting 12 issues of \emph{Variety}.

In order to most closely measure the performance on our final goal (identifying the top $n$ movies by box office per year), we apply the same process of mapping titles to IMDB identifiers for both the gold and extracted sets.
If a movie does not exist in that mapping, we identify it by its lowercased title.  We sum all gross values for the same movie across all tuples to yield a $\langle$ movie, \$ $\rangle$ set and generate a ranking over movies by their total gross.  We do so over the gold annotations and the predicted values to generate two ranked lists.

We measure accuracy with three metrics:

\begin{enumerate}
    \item For all movies that exist in the gold \emph{and} predicted sets, we measure the Spearman rank correlation coefficient $\rho$ between the gross dollar amounts in the two lists. This captures the degree to which the predicted ranks correspond with the true ranks, when we have a gross dollar amount for all movies.
    \item To capture the degree to which the predicted ranks invent movies that do not exist, or fail to extract ones that do, we report the ``Movie F1'' score, where precision is defined as the fraction of predicted movies that exist in gold annotations and recall as the fraction of movies in the gold annotations that exist in the predicted set.  Low precision would mean that models are hallucinating movies that do not exist; low recall would mean that models are failing to identify the ones that do.
    \item While the two metrics above evaluate our final goal (assessing the ranking of movie titles by reported gross), we also evaluate the accuracy of exact tuples (i.e., the degree to which a tuple gets the movie title, theater, city and gross exactly correct).
\end{enumerate}

Table \ref{popeval} reports those metrics over several commercial model variants; each number represents the average of that metric over all \emph{Variety} issues (so that each issue carries equal weight in assessment).

\begin{table}[ht]
\caption{Box office extraction accuracy, with 95\% bootstrap confidence intervals.}
\bigskip
\centering
\begin{tabular}{lcrr}
\toprule
Model & $\rho$ & Movie F1 & Tuple F1 \\
\midrule
Gemini 3 Pro &0.966 {\small[0.953-0.977]} & 0.921 {\small[0.909-0.934]}&0.825 {\small[0.800-0.851]}\\
- (high $\rightarrow$ low thinking level)&0.934 {\small[0.912-0.950]}&0.858 {\small[0.802-0.896]}&0.711 {\small[0.636-0.770]}\\
- (ultrahigh $\rightarrow$ high res)&0.916 {\small[0.887-0.938]} &0.847 {\small[0.800-0.886]}&0.609 {\small[0.533-0.680]} \\
- (pro $\rightarrow$ flash)&0.930 {\small[0.903-0.951]} &0.858 {\small[0.826-0.886]}&0.675 {\small[0.633-0.728]}  \\
\midrule
GPT 5.4 (original res)& 0.916 {\small[0.893-0.937]}&0.809 {\small[0.774-0.845]}&0.684 {\small[0.645-0.725]} \\
- (mini) &0.605 {\small[0.493-0.699]}&0.388 {\small[0.329-0.451]}&0.083 {\small[0.048-0.128]} \\
\bottomrule
\end{tabular}
\label{popeval}
\end{table}

\section{Sample registrations}
\label{registrations}

Figure \ref{fig:bellsreg} illustrates a copyright registration notice for the movie \emph{The Bells of St. Mary's} take from the \emph{Catalog of Copyright Entries, Cumulative Series 1940--1949}; figure \ref{fig:bellsren} shows the copyright registration renewal for that same film exactly 28 years later.  In this case, the renewal references the registration date (6Dec45) and number (L81) of the original.

\begin{figure}[h]
  \centering
  \includegraphics[width=0.7\textwidth]{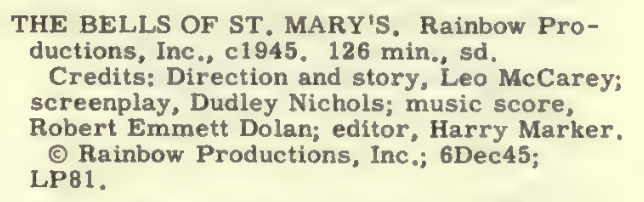}
  \caption{Registration for \emph{Bells of St. Mary's} recorded in the \emph{Catalog of Copyright Entries, Cumulative Series 1940--1949}}
  \label{fig:bellsreg}
\end{figure}

\begin{figure}[h]
  \centering
  \includegraphics[width=0.7\textwidth]{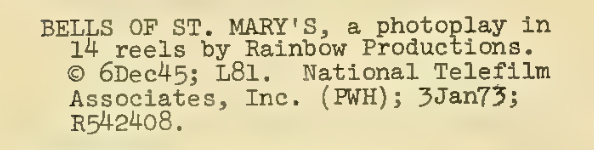}
  \caption{Registration renewal for \emph{Bells of St. Mary's} recorded in the \emph{Catalog of Copyright Entries, Third Series, Volume 27, Parts 12--13, Number 1: Motion Pictures (January--June 1973)}}
  \label{fig:bellsren}
\end{figure}

We use Gemini Pro 2.5 to extract the movie title, registration identifier and copyright date from each entry from the ``Registrations'' section of the CCE---e.g., $\langle$ Bells of St. Mary's, LP81, 6Dec45 $\rangle$---and movie title, original registration number, original copyright date, and renewal number and renewal date from the ``Renewals'' section---e.g., $\langle$ Bells of St. Mary's, LP81, 6Dec45, R542408, 3Jan73 $\rangle$ for volumes from 1957--1978 (corresponding to an earliest original registration date of 1929).  We further extracted all motion picture renewals from the Copyright Public Records System to capture renewals made after 1978.

\section{Memorization}\label{memorization}

As noted in the main text, to identify questions that are highly memorized by models (answerable without access to the movie), we pass all questions through three frontier LLMs---Gemini Pro 3.1, Claude Opus 4.7 and GPT 5.5---with the following prompt: ``Based on your knowledge of the movie \{\texttt{MOVIE}\} (\{\texttt{YEAR}\}), answer the following question.''  All models exhibit similar rates of memorization (Gemini $39.4\%$, Opus $40.7\%$, GPT $38.8\%$).  To mitigate this effect, we subselect questions from the benchmark pool so that the average performance across all models when prompted with the movie title and date alone is approximately $25\%$ (reflecting a true random guess).  The selection process ranks all questions in ascending order by the total number of the three models that correctly answer it from metadata alone, and adds questions sequentially to the benchmark until an average accuracy of $25\%$ is reached.  This yields a total of 628 benchmark questions (discarding 151 from the original pool).  As Table \ref{memcat} illustrates, the exclusion rate varies widely by category. For convenience, we let $\mathcal{D}$ denote the full pool of 779 candidate questions, and $\mathcal{B}$ the post-filter benchmark of 628 questions used in the experiments described \S\ref{sec:experiments}. 

\begin{table}[h!] \small
\centering
\caption{\label{memcat} Exclusion rate by category.}
\bigskip
\begin{tabular}{lrrrr}
\toprule
Category  & Exclusion rate  & $\mathcal{B}$ count & $\mathcal{D}$ count\\
\midrule
Representation        & 0.389 & 22  & 36  \\
Symbolism             & 0.241 & 22  & 29  \\
Temporality           & 0.221 & 53  & 68  \\
Plot                  & 0.209 & 155 & 196 \\
Object Identification & 0.200 & 92  & 115 \\
Character             & 0.167 & 135 & 162 \\
Perspective           & 0.155 & 49  & 58  \\
Setting               & 0.130 & 100 & 115 \\
\midrule
Total                 &        & 628 & 779 \\
\bottomrule
\end{tabular}
\end{table}

\newpage
\section{Frame budget sweep}\label{framebudget}

For the six end-to-end backbones run at multiple frame budgets (GPT-5-mini, Qwen3-VL-8B, and GLM-4.1V-9B-Thinking on the vision--language side; HourLLaVA, VAMBA-Qwen2-VL-7B, and VideoChat-Flash on the long-video side), average accuracy is reported in Table~\ref{tab:frame-sweep} (appendix \ref{framebudget}) for $N \in \{64, 128, 256\}$.
The largest within-backbone gain over the full sweep is GPT-5-mini's $+3.6$ pp from $N{=}64$ to $N{=}256$; the next largest is Qwen3-VL-8B's $+4.5$ pp from $N{=}64$ to $N{=}128$.
Within-backbone CIs overlap heavily across the three budgets, and three of the six backbones (Qwen3-VL-8B, GLM-4.1V-9B-Thinking, VideoChat-Flash, VAMBA-Qwen2-VL-7B in the non-monotone direction) do not improve monotonically with $N$.
The frame-budget effect is an order of magnitude smaller than the audio-access gap shown in \S\ref{sec:experiments-results}, and we therefore use a single canonical budget per sub-paradigm in Table~\ref{tab:acc-final-aligned} ($N{=}64$ for long-video models, $N{=}256$ for vision--language models).

\begin{table}[h!]
  \caption{Frame-budget sweep $N \in \{64, 128, 256\}$ for end-to-end
  vision-language and long-video backbones run at multiple input budgets.}
  \label{tab:frame-sweep}
  \bigskip
  \centering
  \small 
  \setlength{\tabcolsep}{6pt}
  \renewcommand{\arraystretch}{0.95}
  \begin{tabular}{lccc}
    \toprule
    Backbone & $N{=}64$ & $N{=}128$ & $N{=}256$ \\
    \midrule
    \multicolumn{4}{l}{\textit{End-to-end vision-language models}} \\
    \midrule
    GPT-5-mini & \val{30.6}{27.2--33.9} & \val{32.8}{29.3--36.3} & \val{34.2}{30.7--37.7} \\
    Qwen3-VL-8B & \val{25.3}{22.0--28.5} & \val{29.8}{26.4--33.1} & \val{28.8}{25.5--32.2} \\
    GLM-4.1V-9B-Thinking & \val{25.8}{22.6--29.0} & \val{25.6}{22.5--28.8} & \val{24.2}{21.2--27.4} \\
    \midrule
    \multicolumn{4}{l}{\textit{End-to-end long-video models}} \\
    \midrule
    HourLLaVA & \val{23.6}{20.4--26.8} & \val{24.7}{21.5--27.9} & \val{26.3}{23.1--29.5} \\
    VAMBA-Qwen2-VL-7B & \val{23.7}{20.5--26.9} & \val{25.2}{22.0--28.5} & \val{23.2}{20.1--26.4} \\
    VideoChat-Flash & \val{27.4}{24.0--30.7} & \val{27.2}{23.9--30.6} & \val{27.1}{23.7--30.4} \\
    \bottomrule
  \end{tabular}
\end{table}

\section{Movies in Benchmark}

{\small
\begin{longtable}{>{\raggedright\arraybackslash}p{3cm} c >{\raggedright\arraybackslash}p{3cm} >{\raggedright\arraybackslash}p{2.5cm} >{\raggedright\arraybackslash}p{2.5cm}}
\toprule

\textbf{Title} & \textbf{Year} & \textbf{IMDB Genres} & \textbf{Director} & \textbf{Production Company} \\
\midrule
\endfirsthead

\multicolumn{5}{c}{\textit{(continued from previous page)}} \\[4pt]
\toprule
\textbf{Title} & \textbf{Year} & \textbf{IMDB Genres} & \textbf{Director} & \textbf{Production Company} \\
\midrule
\endhead

\midrule
\multicolumn{5}{r}{\textit{Continued on next page}} \\
\endfoot

\bottomrule
\endlastfoot

McLintock!                       & 1963 & Comedy, Western                & Andrew V. McLaglen  & Batjac Productions \\
Beneath the 12-Mile Reef         & 1953 & Adventure, Drama, Romance      & Robert D. Webb      & 20th Century Fox \\
Cyrano de Bergerac               & 1951 & Adventure, Drama, Romance      & Michael Gordon      & Stanley Kramer Productions \\
Go for Broke!                    & 1951 & Drama, History, War            & Robert Pirosh       & Loew's \\
Royal Wedding                    & 1951 & Comedy, Musical, Romance       & Stanley Donen       & Loew's \\
Three Guys Named Mike            & 1951 & Comedy, Romance                & Charles Walters     & Metro-Goldwyn-Mayer (MGM) \\
The Inspector General            & 1949 & Comedy, Musical, Romance       & Henry Koster        & Warner Bros. \\
Tulsa                            & 1949 & Drama, Western                 & Stuart Heisler      & Walter Wanger Productions \\
He Walked by Night               & 1948 & Crime, Drama, Film-Noir        & Alfred L. Werker    & Bryan Foy Productions \\
My Favorite Brunette             & 1947 & Comedy, Crime, Mystery         & Elliott Nugent      & Hope Enterprises \\
The Perils of Pauline            & 1947 & Drama, Romance                 & George Marshall     & Paramount Pictures \\
Smash Up: The Story of a Woman   & 1947 & Comedy, Crime, Drama           & Stuart Heisler      & Walter Wanger Productions \\
Till the Clouds Roll By          & 1947 & Biography, Musical             & Richard Whorf       & Metro-Goldwyn-Mayer (MGM) \\
Angel on My Shoulder             & 1946 & Adventure, Comedy, Fantasy     & Archie Mayo         & Charles R. Rogers Productions \\
The Strange Love of Martha Ivers & 1946 & Drama, Film-Noir, Romance      & Lewis Milestone     & Hal Wallis Productions \\
The Stranger                     & 1946 & Crime, Drama, Film-Noir        & Orson Welles        & International Pictures, The Haig Corporation \\
Blood on the Sun                 & 1945 & Drama, Romance, Thriller       & Frank Lloyd         & William Cagney Productions \\
Captain Kidd                     & 1945 & Adventure, Biography, Drama    & Rowland V. Lee      & Benedict Bogeaus Production \\
The Stork Club                   & 1945 & Comedy, Musical, Romance       & Hal Walker          & B.G. DeSylva Productions Inc. \\
Stage Door Canteen               & 1943 & Comedy, Music, Romance         & Frank Borzage       & Sol Lesser Productions \\
Rudyard Kipling's Jungle Book    & 1942 & Action, Adventure, Family      & Zoltan Korda        & Alexander Korda Films \\
Billy the Kid in Santa Fe        & 1941 & Drama, Western                 & Sam Newfield        & Sigmund Neufeld Productions \\
Meet John Doe                    & 1941 & Comedy, Drama, Romance         & Frank Capra         & Frank Capra Productions \\
Second Chorus                    & 1941 & Comedy, Musical, Romance       & H.C. Potter         & Boris Morros Productions \\
His Girl Friday                  & 1940 & Comedy, Drama, Romance         & Howard Hawks        & Columbia Pictures \\
Gulliver's Travels               & 1939 & Adventure, Animation, Comedy   & Dave Fleischer      & Fleischer Studios \\
The Little Princess              & 1939 & Comedy, Drama, Family          & Walter Lang         & 20th Century Fox \\
Love Affair                      & 1939 & Comedy, Drama, Romance         & Leo McCarey         & RKO Radio Pictures \\
Made for Each Other              & 1939 & Comedy, Drama, Romance         & John Cromwell       & Selznick International Pictures \\
Nurse Edith Cavell               & 1939 & Biography, Drama, War          & Herbert Wilcox      & Imperadio Pictures Ltd. \\
Letter of Introduction           & 1938 & Comedy, Drama, Mystery         & John M. Stahl       & Universal Pictures \\
A Star Is Born                   & 1937 & Drama, Romance                 & William A. Wellman  & Selznick International Pictures \\
Swing High, Swing Low            & 1937 & Comedy, Drama, Musical         & Mitchell Leisen     & Paramount Pictures \\
Little Lord Fauntleroy           & 1936 & Drama, Family                  & John Cromwell       & Selznick International Pictures \\
Becky Sharp                      & 1935 & Drama, Romance, War            & Rouben Mamoulian    & Pioneer Pictures Corporation \\
Of Human Bondage                 & 1934 & Drama, Film-Noir, Romance      & John Cromwell       & RKO Radio Pictures \\
A Farewell to Arms               & 1932 & Drama, Romance, War            & Frank Borzage       & Paramount Pictures \\
Bird of Paradise                 & 1932 & Adventure, Drama, Romance      & King Vidor          & RKO Radio Pictures \\
Rain                             & 1932 & Drama                          & Lewis Milestone     & Feature Productions \\
The Front Page                   & 1931 & Comedy, Crime, Drama           & Lewis Milestone     & The Caddo Company \\
Parlor, Bedroom and Bath         & 1931 & Comedy                         & Edward Sedgwick     & Metro-Goldwyn-Mayer (MGM) \\
Street Scene                     & 1931 & Drama, Romance                 & King Vidor          & The Samuel Goldwyn Company, Feature Productions \\
All Quiet on the Western Front   & 1930 & Drama, War                     & Lewis Milestone     & Universal Pictures \\
Animal Crackers                  & 1930 & Comedy, Family, Musical        & Victor Heerman      & Paramount Pictures \\
Anybody's Woman                  & 1930 & Drama, Romance                 & Dorothy Arzner      & Paramount Pictures \\
The Big House                    & 1930 & Crime, Drama, Thriller         & George W. Hill      & Metro-Goldwyn-Mayer (MGM), Cosmopolitan Productions \\
The Big Trail                    & 1930 & Adventure, Drama, Romance      & Raoul Walsh         & Fox Film Corporation \\
Check and Double Check           & 1930 & Comedy                         & Melville W. Brown   & RKO Radio Pictures \\
The Divorcee                     & 1930 & Drama, Romance                 & Robert Z. Leonard   & Metro-Goldwyn-Mayer (MGM) \\
Feet First                       & 1930 & Adventure, Comedy, Family      & Clyde Bruckman      & The Harold Lloyd Corporation \\
Min and Bill                     & 1930 & Comedy, Drama                  & George W. Hill      & Metro-Goldwyn-Mayer (MGM) \\
Reaching for the Moon            & 1930 & Comedy, Romance                & Edmund Goulding     & Feature Productions \\
Song o' My Heart                 & 1930 & Drama, Music, Romance          & Frank Borzage       & Fox Film Corporation \\
Bulldog Drummond                 & 1929 & Crime, Drama, Mystery          & F. Richard Jones    & The Samuel Goldwyn Company \\
The Canary Murder Case           & 1929 & Crime, Drama, Mystery          & Malcolm St. Clair   & Paramount Pictures \\
Coquette                         & 1929 & Drama, Romance                 & Sam Taylor          & Pickford Corporation \\
Happy Days                       & 1929 & Comedy, Musical, Romance       & Benjamin Stoloff    & Fox Film Corporation \\
Sally                            & 1929 & Musical                        & John Francis Dillon & First National Pictures \\
The Trespasser                   & 1929 & Drama, Romance                 & Edmund Goulding     & Gloria Productions \\
Weary River                      & 1929 & Drama, Romance                 & Frank Lloyd         & First National Pictures \\
The Wild Party                   & 1929 & Comedy, Drama, Romance         & Dorothy Arzner      & Paramount Pictures \\

\end{longtable}
}

\end{document}